\documentclass[letterpaper, 10pt, conference]{ieeeconf}
\IEEEoverridecommandlockouts
\usepackage{amsmath, amssymb}
\usepackage{graphicx}
\usepackage{booktabs}
\usepackage{multirow}
\usepackage{siunitx}
\usepackage{caption}
\usepackage{subcaption}
\usepackage{algorithm}
\usepackage{algpseudocode}
\usepackage{cite}
\usepackage{hyperref}
\usepackage[utf8]{inputenc} % Allows UTF-8 characters
\usepackage[T1]{fontenc}    % Standard font encoding
\usepackage{textcomp}       % For symbols like km/s
\usepackage{tabularx}
\usepackage{booktabs}

\title{\LARGE \bf
Evaluating Robustness and Adaptability in Learning-Based Mission Planning for Active Debris Removal
}

\author{Agni Bandyopadhyay$^{1}$ and Günther Waxenegger-Wilfing$^{2}$% <-this % stops a space
\thanks{$^{1}$Agni Bandyopadhyay is pursuing his Doctoral degree with the Faculty of Mathematics and Computer Science,
        Julius-Maximilians-Universität Würzburg, Sanderring 2, 97070 Würzburg, Germany
        {\tt\small agni.bandyopadhyay@uni-wuerzburg.de}}%
\thanks{$^{2}$Günther Waxenegger-Wilfing is a Professor with the Faculty of Mathematics and Computer Science,
        Julius-Maximilians-Universität Würzburg, Sanderring 2, 97070 Würzburg, Germany
        {\tt\small guenther.waxenegger@uni-wuerzburg.de}}%
}
\begin{document}

\maketitle
\thispagestyle{empty}
\pagestyle{empty}

%%%%%%%%%%%%%%%%%%%%%%%%%%%%%%%%%%%%%%%%%%%%%%%%%%%%%%%%%%%%%%%%%%%%%%%%%%%%%%%%
\begin{abstract}
Autonomous mission planning for Active Debris Removal (ADR) must balance efficiency, adaptability, and strict feasibility constraints on fuel and mission duration. This work compares three planners for the constrained multi-debris rendezvous problem in Low Earth Orbit: a \textbf{nominal Masked Proximal Policy Optimization (PPO)} policy trained under fixed mission parameters, a \textbf{domain-randomized Masked PPO} policy trained across varying mission constraints for improved robustness, and a \textbf{plain Monte Carlo Tree Search (MCTS)} baseline. Evaluations are conducted in a high-fidelity orbital simulation with refueling, realistic transfer dynamics, and randomized debris fields across 300 test cases in nominal, reduced fuel, and reduced mission time scenarios. Results show that nominal PPO achieves top performance when conditions match training but degrades sharply under distributional shift, while domain-randomized PPO exhibits improved adaptability with only moderate loss in nominal performance. MCTS consistently handles constraint changes best due to online replanning but incurs orders-of-magnitude higher computation time. The findings underline a trade-off between the speed of learned policies and the adaptability of search-based methods, and suggest that combining training-time diversity with online planning could be a promising path for future resilient ADR mission planners.
\end{abstract}

\section{Introduction}

The exponential growth of artificial satellites, mega-constellations, and orbital debris in Low Earth Orbit (LEO) has heightened global concern regarding the long-term sustainability of space operations \cite{esa2023}. As orbital density increases, the probability of cascading collisions—commonly referred to as the Kessler Syndrome—becomes more than a theoretical threat \cite{kessler1978}. This makes future Active Debris Removal (ADR) missions increasingly critical to preserving the safety and functionality of essential space infrastructure \cite{nasa2023}.

A key challenge in autonomous ADR lies in \emph{mission planning under hard operational constraints}. Given limited fuel budgets and strict mission durations, future autonomous ADR spacecraft will need to determine which debris objects to rendezvous with and in what sequence, potentially incorporating capabilities such as mid-mission refueling and complex orbital transfers. This problem is naturally formulated as a sequential decision-making task under feasibility constraints \cite{tsp2019}, where only certain actions remain valid at each step depending on dynamic resource availability and mission deadlines.

Two broad classes of autonomous decision-making have emerged for such problems: \emph{learned policy} approaches and \emph{online search-based} planning. Reinforcement learning (RL) methods such as Proximal Policy Optimization (PPO) \cite{schulman2017ppo} are attractive for onboard deployment because, once trained, the policy can generate actions with very low latency. However, these policies can degrade in performance when deployment conditions differ from the training distribution, a phenomenon widely known as distributional shift \cite{quinonero2009datasetshift,shimodaira2000improving}. In ADR, this may occur when available $\Delta v$, mission duration, or another mission parameter differs from those assumed during training.

In contrast, online planning methods such as Monte Carlo Tree Search (MCTS) \cite{kocsis2006bandit} adapt their decision-making at each step by simulating future outcomes from the current state. This allows them to handle unexpected mission constraints more gracefully. However, the computational cost of repeated simulations makes such methods difficult to run in real time on resource-limited spacecraft hardware, especially in large or heavily constrained action spaces.

In this work, we investigate the \emph{robustness and adaptability} of learned-policy and online planning strategies for multi-debris rendezvous in LEO. We evaluate three planners: a nominal PPO policy trained under fixed mission parameters, a domain-randomized PPO policy trained across varying $\Delta v$ and mission durations to enhance robustness, and a plain MCTS baseline. By testing these planners in nominal, reduced fuel, and reduced mission time scenarios, we characterize how each approach adapts its decision-making under changing operational constraints.

The results provide practical guidance for ADR mission planning: when to rely on fast but potentially brittle learned policies, when to use robust but computationally heavy online search, and how training diversity can bridge the gap between efficiency and adaptability for autonomous space operations.

\section{Related Work}

Mission planning for Active Debris Removal (ADR) has been widely explored through combinatorial optimization and trajectory design. Classical approaches often employ graph-based heuristics or mixed-integer programming to solve the debris collection problem \cite{liu2015optimal, bombrun2013multi}. While effective under static assumptions, these methods are typically brittle when mission parameters change during evaluation and require extensive hand-tuning to remain applicable across diverse scenarios.

More recently, reinforcement learning (RL) has emerged as a promising alternative due to its ability to learn complex decision-making strategies directly from simulated interactions and to provide fast inference at runtime. Deep RL methods have been applied to spacecraft guidance, proximity operations, and orbital transfers \cite{chi2021deep, gaudet2020deep}. In ADR, several studies have demonstrated the feasibility of using RL to plan multi-target debris collection missions while minimizing fuel consumption \cite{bandyopadhyay2024isparo, zhao2021space}. Our previous work introduced a masked PPO agent that incorporated feasibility constraints such as refueling, $\Delta v$ limits, and collision avoidance into its decision-making process \cite{bandyopadhyay2025adaptive}, focusing on performance in a nominal mission configuration \cite{bandyopadhyay2025camsat}.

A recurring challenge for learned policies is their sensitivity to \emph{distributional shift}—the degradation in performance that occurs when deployment conditions differ from those seen during training \cite{quinonero2009datasetshift, shimodaira2000improving}. In ADR, this can occur when mission parameters change between training and deployment, with two critical examples being variations in available $\Delta v$ (fuel budget) and total mission duration. 

In contrast to learned policies, online planning methods such as Monte Carlo Tree Search (MCTS) \cite{kocsis2006bandit} adapt their decision-making at every step by simulating the outcomes of feasible actions from the current state. MCTS has been successfully applied in robotics and space systems for fault recovery, autonomous navigation, and path planning under uncertainty \cite{cheng2022adaptive}. This dynamic replanning makes MCTS inherently more robust to changes in mission constraints, although its computational expense and simulation overhead limit its practicality for real-time onboard execution, particularly in environments with large or heavily masked action spaces.

While both RL-based and MCTS-based approaches have been investigated for ADR, there is limited work directly analyzing how these methods adapt to \emph{changing operational constraints}. This work addresses that gap by evaluating a nominal PPO policy, a domain-randomized PPO policy trained for robustness, and a plain MCTS planner across nominal, reduced fuel, and reduced mission time scenarios, providing insight into the trade-offs between decision speed, adaptability, and performance in constrained ADR missions.

\section{Problem Statement}

We model autonomous Active Debris Removal (ADR) in Low Earth Orbit (LEO) as a constrained sequential decision-making problem. The service spacecraft operates in a 700--800 km altitude band with low eccentricity ($e < 0.01$), starting each mission docked at a refueling station in a near-circular 700 km orbit. A set of 50 debris objects, each with unique orbital parameters, is generated at the start of each episode.

At each decision step, the spacecraft selects either:
\begin{enumerate}
    \item A transfer to an unvisited debris object, or
    \item A transfer to the refueling station.
\end{enumerate}
Feasible actions are determined by remaining $\Delta v$, remaining mission time, and debris visitation status.

Two global mission constraints apply:
\begin{itemize}
    \item Total mission duration limit,
    \item Cumulative $\Delta v$ budget, replenishable only via refueling.
\end{itemize}

\begin{figure}[!t]
    \centering
    \includegraphics[width=\linewidth]{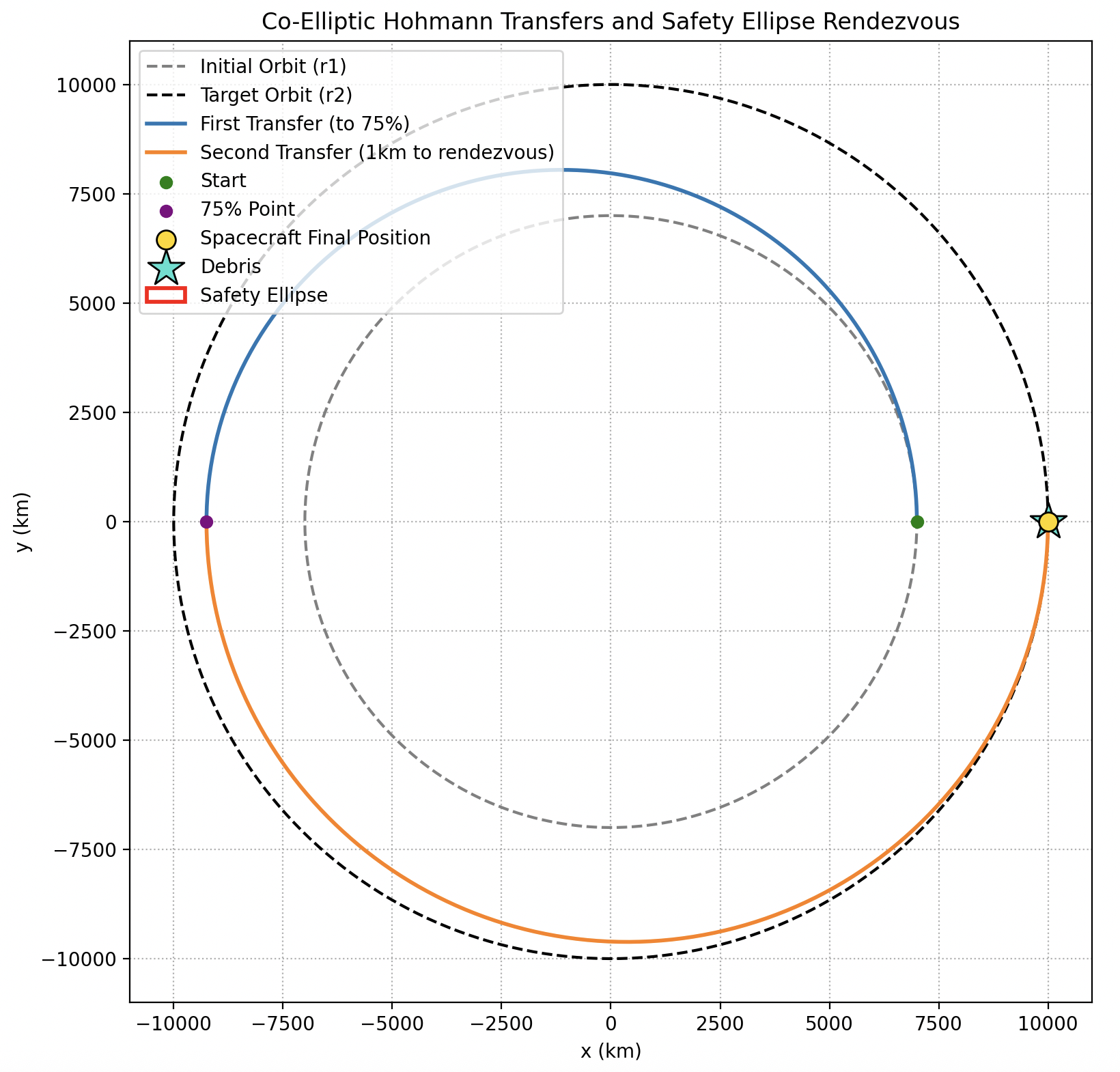}
    \caption{Overview of the co-elliptic Hohmann transfer sequence, showing the initial and target orbits, transfer maneuvers, and the terminal safety ellipse maneuver at rendezvous.}
    \label{fig:coelliptic}
\end{figure}

\begin{figure}[!t]
    \centering
    \includegraphics[width=\linewidth]{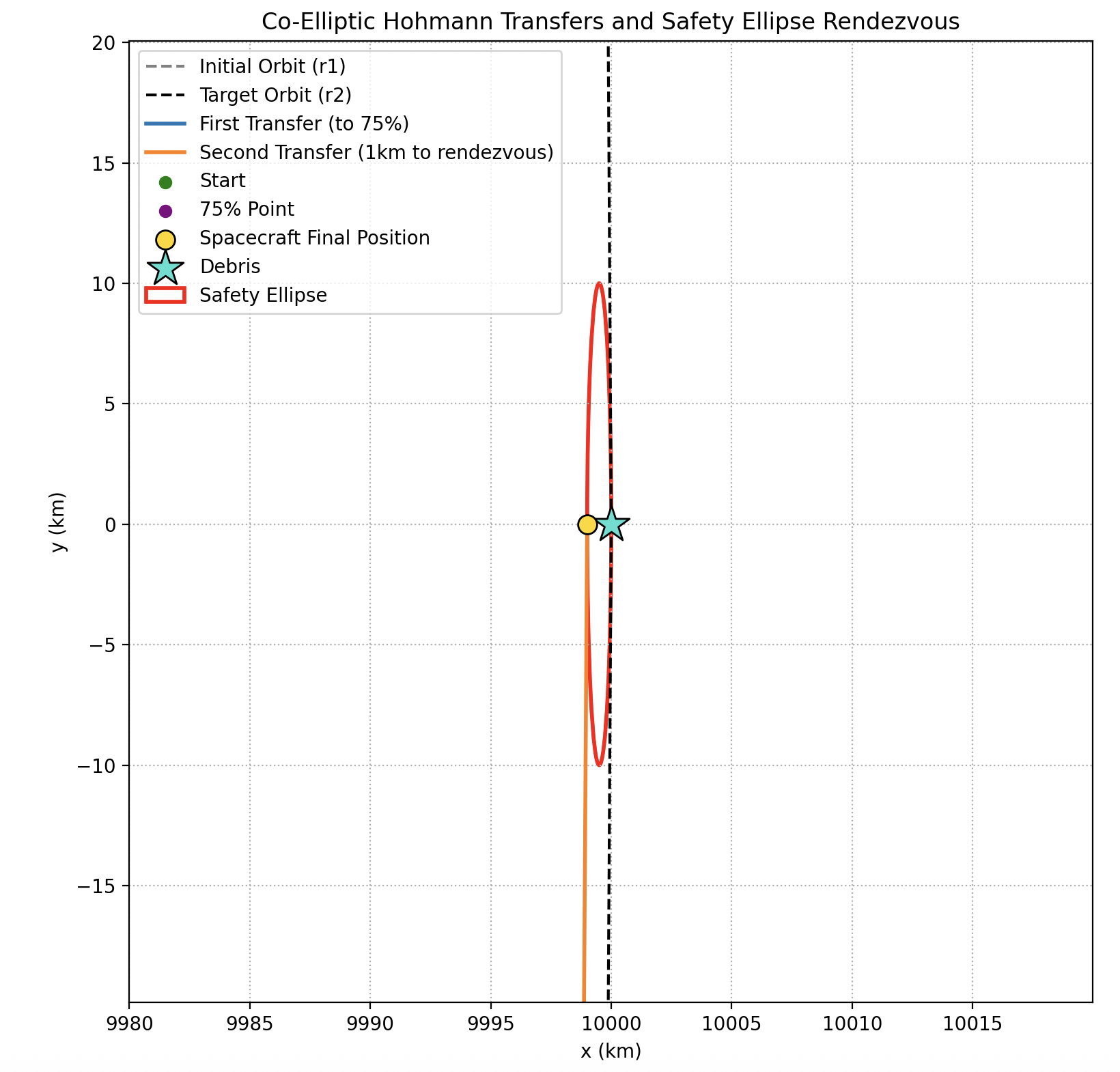}
    \caption{Zoomed-in view of the terminal approach phase, highlighting the safety ellipse maneuver for controlled and safe rendezvous with the target debris.}
    \label{fig:safetyellipse}
\end{figure}

Transfers between targets follow a co-elliptic Hohmann maneuver strategy \cite{nasaatm2010}, as illustrated in Figure~\ref{fig:coelliptic}:
\begin{enumerate}
    \item A Hohmann transfer to a co-elliptic orbit, placing the chaser approximately 75\% of the way to the target,
    \item A second burn to reduce separation to within 1 km,
    \item A terminal safety ellipse maneuver for controlled final approach \cite{alonso2009collision}, shown in Figure~\ref{fig:safetyellipse}.
\end{enumerate}

The refueling process restores the $\Delta v$ budget to its maximum, but consumes mission time and maneuvering cost. Debris objects are considered ``visited'' upon successful rendezvous and are masked out of future decisions. Each episode terminates when time or $\Delta v$ is exhausted, mission time is completed or when no feasible actions remain.

We evaluate planning strategies across three scenarios:
\begin{itemize}
    \item \textbf{Nominal:} 7-day mission duration, 3 km/s $\Delta v$ budget.
    \item \textbf{Reduced fuel:} $\Delta v$ budget reduced to 1 km/s.
    \item \textbf{Reduced mission time:} mission duration reduced to 3 days.
\end{itemize}

The objective is to maximize the number of successful debris rendezvous while satisfying all mission constraints.

\section{Methods}

We evaluate two decision-making algorithms for constrained multi-debris rendezvous in Low Earth Orbit: \textit{Masked Proximal Policy Optimization (PPO)} and \textit{Monte Carlo Tree Search (MCTS)}. Both methods operate over the same discrete action space \( A \), where each action corresponds to transferring to an unvisited debris object or to the refueling station. Feasibility constraints—based on remaining $\Delta v$, remaining mission time, and debris visitation status—are enforced through action masking.

\subsection{Masked Proximal Policy Optimization}

We use the Maskable Proximal Policy Optimization (MaskablePPO) algorithm from Stable-Baselines3-Contrib \cite{stable-baselines3}, which extends PPO \cite{schulman2017ppo} with native support for action masking during training. The binary action mask \( m(s) \in \{0,1\}^A \) identifies feasible actions in a given state \( s \), ensuring that invalid transfers never influence policy updates or are sampled for execution. This is critical in ADR mission planning, where infeasible maneuvers must be strictly avoided.

The policy is optimized using the clipped surrogate loss \cite{schulman2017ppo}:
\[
\mathcal{L}_{\text{CLIP}} = -\min \left( r_t \hat{A}_t,\ \text{clip}(r_t,\ 1 - \epsilon,\ 1 + \epsilon) \hat{A}_t \right),
\]
where \( r_t \) is the probability ratio between new and old policies, and \( \hat{A}_t \) is the estimated advantage. The hyperparameter \( \epsilon \) limits policy update magnitude for stability.

The state vector includes:
\begin{itemize}
    \item Orbital elements of the chaser spacecraft,
    \item Relative positions and velocities of debris,
    \item Remaining $\Delta v$ and elapsed mission time,
    \item Refueling status,
    \item A visitation mask for debris already serviced.
\end{itemize}

Rewards are shaped to encourage debris removal while penalizing excessive refueling:
\begin{itemize}
    \item +1 for each successful rendezvous,
    \item –0.5 for each refueling trip.
\end{itemize}

The reward ratio was empirically chosen to balance the trade-off between aggressive debris collection and excessive refueling. A higher penalty discouraged refueling and led to premature mission termination, while a lower penalty encouraged fuel-inefficient hopping between debris. The selected ratio achieved stable learning and overall better mission behavior.

\subsection{Monte Carlo Tree Search}

We implement a plain Monte Carlo Tree Search (MCTS) algorithm using the Upper Confidence bounds applied to Trees (UCT) selection rule \cite{russell2021artificial}:
\[
\text{UCT}(s,a) = Q(s,a) + c_{\text{uct}} \cdot \sqrt{\frac{\log N(s)}{1 + N(s,a)}},
\]
where \( Q(s,a) \) is the mean return, \( N(s,a) \) is the visit count for action \( a \), and \( N(s) \) is the total visits to node \( s \). The constant \( c_{\text{uct}} \) balances exploration and exploitation.

The search process consists of:
\begin{enumerate}
    \item \textbf{Selection:} choose actions using UCT until reaching a leaf node.
    \item \textbf{Expansion:} add child nodes for feasible actions (\( m_a(s) = 1 \)).
    \item \textbf{Simulation:} clone the environment and simulate with a uniform random policy over valid actions until a terminal condition (time/fuel exhaustion or no valid actions).
    \item \textbf{Backpropagation:} update \( Q(s,a) \) and visit counts along the path.
\end{enumerate}

The same action-masking mechanism is used as in PPO, ensuring consistent feasibility checks. Each decision step uses a fixed simulation budget, with rollouts depth-limited to reduce computational cost.

While MCTS dynamically replans at each step and can adapt to mission changes without retraining, its heavy reliance on repeated simulation makes it significantly slower than PPO, limiting its suitability for real-time onboard execution in resource-constrained spacecraft.

\begin{figure*}[!t]
    \centering
    \includegraphics[width=\textwidth]{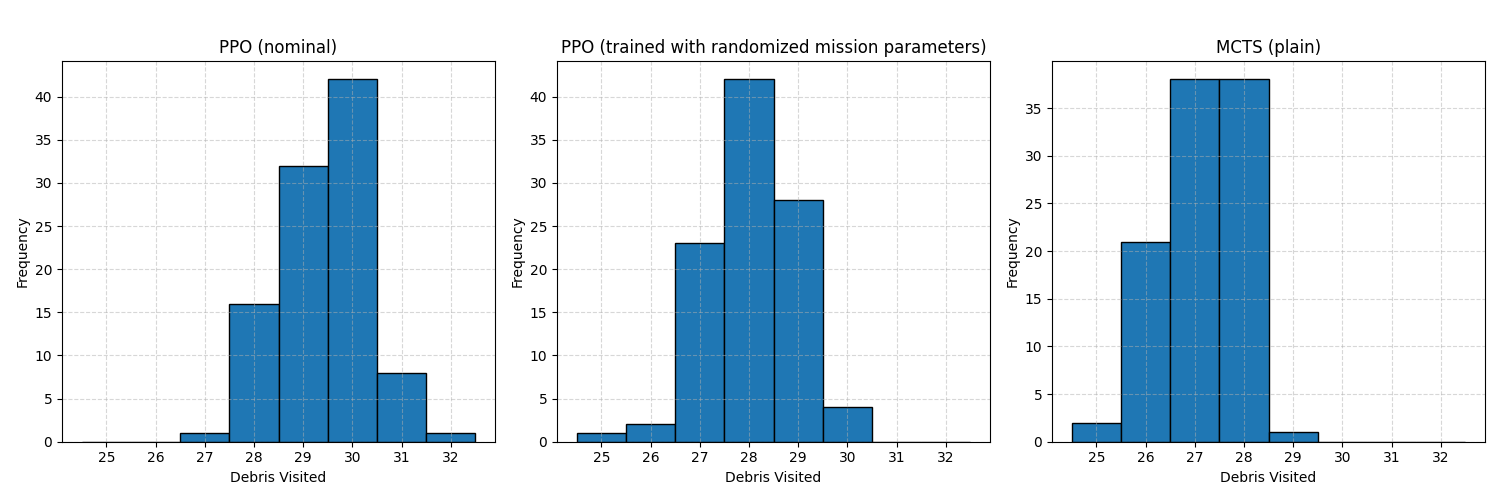}
    \caption{Debris visited under nominal mission conditions (3 km/s dv, 7-day duration) for PPO (nominal), PPO (domain-randomized), and MCTS.}
    \label{fig:nominal}
\end{figure*}

\begin{figure*}[!t]
    \centering
    \includegraphics[width=\textwidth]{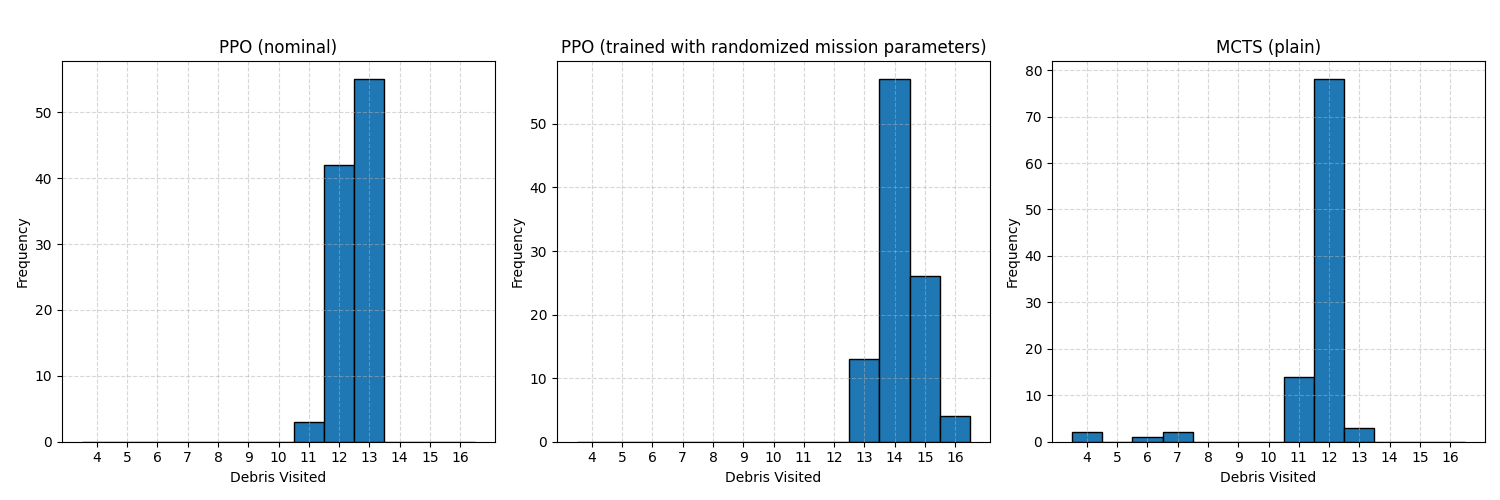}
    \caption{Debris visited under reduced mission duration (3 days) for PPO (nominal), PPO (domain-randomized), and MCTS.}
    \label{fig:time}
\end{figure*}

\section{Experimental Setup}

All experiments are conducted in the custom \texttt{SpaceDebrisStressTestEnv}, a Gymnasium-compatible environment simulating multi-target debris rendezvous in LEO with dynamic feasibility constraints. Each episode initializes 50 randomly generated debris objects in near-circular orbits between 700--800~km altitude and inclination angles between 94°--98°.The chaser spacecraft and the refueling station are both initialized in a 700~km, 96° almost circular orbit, with the chaser tasked with maximizing debris collection within the mission constraints.

\subsection{Evaluated Planners}

\paragraph{Nominal PPO} The first PPO configuration is trained using the MaskablePPO implementation from Stable-Baselines3-Contrib \cite{stable-baselines3}, with action masking enabled to enforce feasibility. The policy network is a multilayer perceptron (MLP) with two hidden layers of size 256, trained for 1 million timesteps using the Adam optimizer. Default PPO hyperparameters are used, with a learning rate of \( 3 \times 10^{-5} \). Training is performed under fixed mission parameters matching the nominal scenario.

\paragraph{Domain-Randomized PPO} The second PPO configuration is trained identically, except that the mission parameters (total mission duration and $\Delta v$ budget) are randomized at the start of each episode within predefined ranges and was trained for 5.5 million timesteps. This domain randomization exposes the policy to both nominal and constrained regimes during training, aiming to improve robustness under changing mission conditions.

\paragraph{MCTS} The MCTS agent is implemented using the plain Upper Confidence bounds for Trees (UCT) algorithm \cite{russell2021artificial}, running for 200 simulations per decision step with the exploration constant \( c_{\text{uct}} = 1.5 \). The simulation budget of 200 rollouts per decision step was determined empirically by observing performance plateaus; increasing beyond this point yielded negligible improvements while significantly increasing computation time. The MCTS agent leverages the same dynamic action masking via the \texttt{valid\_action\_mask()} method as PPO, ensuring identical feasibility handling.

\subsection{Test Scenarios}

To evaluate robustness under varying operational constraints, we define three mission scenarios:
\begin{enumerate}
    \item \textbf{Nominal:} 7-day mission duration, 3~km/s $\Delta v$ budget (matches Nominal PPO training).
    \item \textbf{Reduced fuel:} $\Delta v$ budget reduced to 1~km/s.
    \item \textbf{Reduced mission time:} mission duration reduced to 3~days.
\end{enumerate}

Each method is evaluated on 100 independent test cases per scenario. Debris sets are randomly generated per run, and performance is measured in terms of the number of successful rendezvous.

\subsection{Execution Environment}

All experiments are conducted on a MacBook Pro with an Apple M1 Max chip using CPU-only inference.

\begin{figure*}[!t]
    \centering
    \includegraphics[width=\textwidth]{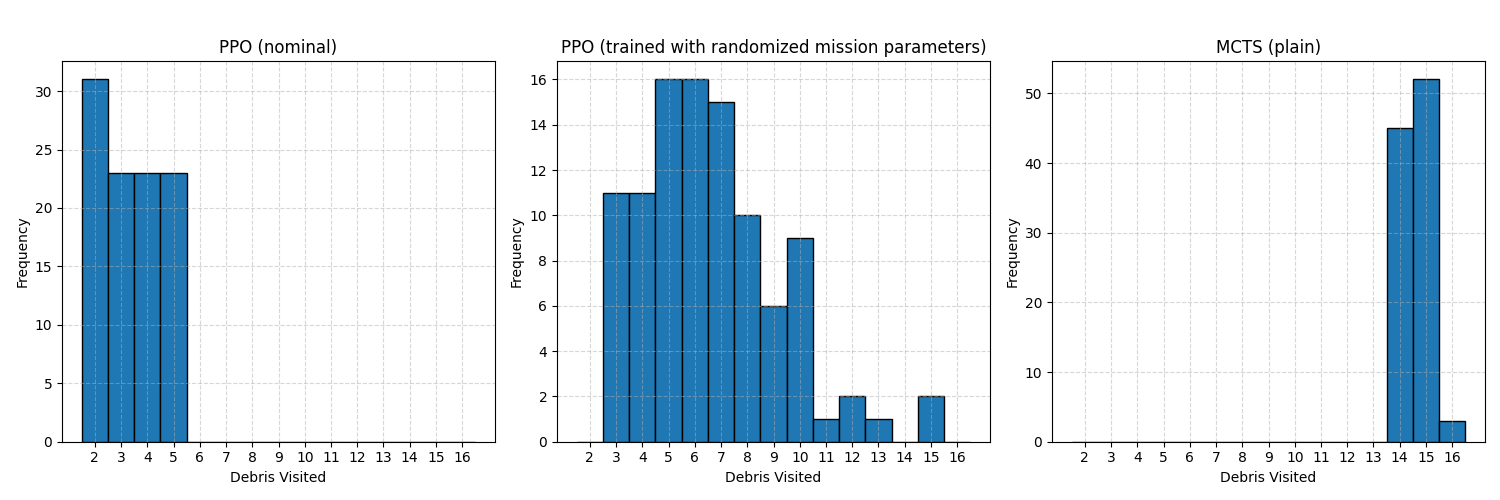}
    \caption{Debris visited under reduced dv budget (1 km/s) for PPO (nominal), PPO (domain-randomized), and MCTS.}
    \label{fig:dv}
\end{figure*}

\section{Results and Discussion}

This section presents a comparative evaluation of three planning strategies: \textit{Masked PPO} (nominal training), \textit{Masked PPO} trained with domain-randomized mission parameters, and \textit{plain MCTS}, across three mission scenarios—nominal, reduced mission duration, and reduced $\Delta v$. Each method was tested on 100 independently generated debris fields per scenario.

\subsection*{Nominal Scenario}

Figure~\ref{fig:nominal} shows the debris removal performance under nominal mission conditions (3~km/s $\Delta v$, 7-day duration). Both PPO variants achieve high and consistent performance, visiting nearly 30 debris objects on average. However the nominal one performs slightly better as it was able to specialize for that particular scenario. MCTS, while slightly lower in average performance, remains competitive, highlighting its ability to find effective solutions even without prior training.

\subsection*{Reduced Mission Duration Scenario}

In the reduced-time scenario (Figure~\ref{fig:time}), where the mission duration is shortened to 3~days, all methods see a drop in performance. Domain-randomized PPO shows the best adaptability, outperforming both nominal PPO and MCTS in average debris count. MCTS remains competitive but is slightly behind due to limited time for exploration-based planning.

\subsection*{Reduced $\Delta v$ Scenario}

The reduced-fuel case (Figure~\ref{fig:dv}), with only 1~km/s maneuvering budget, exposes the limitations of nominal PPO under distributional shift—it often depletes fuel early, achieving poor coverage. Domain-randomized PPO adapts far better, visiting over twice as many debris objects as nominal PPO, but still falls short of MCTS. Thanks to its online replanning at every step, MCTS achieves the highest performance under this severe constraint.

\begin{table}[!t]
\centering
\caption{Debris Removal Statistics Across 100 Test Cases}
\label{tab:summary}
\begin{tabular}{lcccc}
\toprule
\multirow{2}{*}{\textbf{Scenario}} & \multirow{2}{*}{\textbf{Method}} & \multicolumn{3}{c}{\textbf{Debris Visited}} \\
\cmidrule(lr){3-5}
& & \textbf{Min} & \textbf{Max} & \textbf{Avg $\pm$ Std} \\
\midrule
Nominal & PPO (nom.) & 27 & 31 & 29.1 $\pm$ 1.1 \\
        & PPO (rand.) & 25 & 30 & 28.2 $\pm$ 1.2 \\
        & MCTS & 25 & 28 & 27.1 $\pm$ 0.9 \\
\midrule
Time-limited & PPO (nom.) & 11 & 13 & 12.6 $\pm$ 0.6 \\
             & PPO (rand.) & 13 & 16 & 14.1 $\pm$ 0.7 \\
             & MCTS & 11 & 12 & 11.9 $\pm$ 0.3 \\
\midrule
$\Delta v$-limited & PPO (nom.) & 2 & 5 & 3.2 $\pm$ 0.9 \\
                   & PPO (rand.) & 4 & 15 & 8.1 $\pm$ 2.9 \\
                   & MCTS & 14 & 16 & 15.0 $\pm$ 0.4 \\

\bottomrule
\end{tabular}
\end{table}

\subsection*{Computation Time}

\begin{figure*}[!t]
    \centering
    \includegraphics[width=\textwidth]{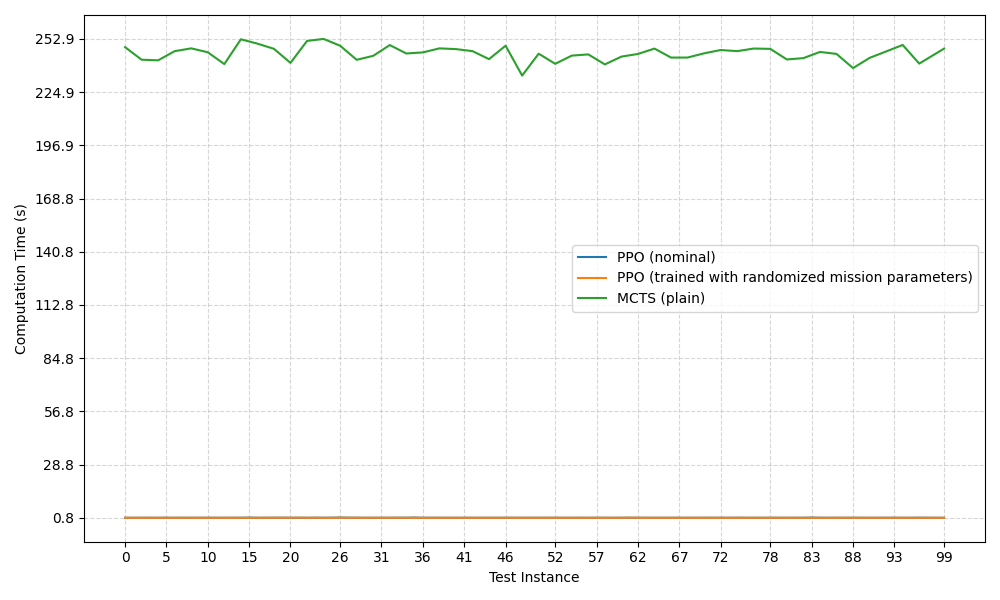}
    \caption{Average execution time per test case (s) for each method across all scenarios. The two PPO curves overlap due to identical inference times.}
    \label{fig:comptime}
\end{figure*}

Figure~\ref{fig:comptime} compares computation time per test instance. Both PPO variants require under 1~second per episode on average, making them indistinguishable in the plot—their curves overlap. In contrast, MCTS averages over 4~minutes per case, dominated by repeated environment cloning and rollouts.

\subsection*{Summary}

Overall, nominal PPO excels in scenarios matching its training distribution but suffers under shifted constraints. Domain-randomized PPO mitigates this drop, showing improved adaptability in reduced fuel and time scenarios. MCTS consistently handles constraint changes best due to its online nature, though at a substantial computational cost. These results underline the trade-off between the fast inference of learned policies and the adaptability of search-based methods in ADR mission planning.

\section{Conclusion and Future Work}

This work presented a comparative evaluation of three decision-making approaches for the constrained multi-debris rendezvous planning problem in Low Earth Orbit: \textit{Masked Proximal Policy Optimization (PPO)}, \textit{domain-randomized Masked PPO}, and \textit{plain Monte Carlo Tree Search (MCTS)}. 

Under nominal mission conditions, both PPO variants achieved similarly high performance and identical sub-second inference times, making them equally well-suited for real-time onboard decision-making. However, their behaviour diverged under distributional shift. When mission constraints were altered—such as reduced $\Delta v$ or shortened mission duration—the nominal-trained PPO suffered significant performance degradation, particularly in the $\Delta v$-limited scenario. The domain-randomized PPO, trained across varying mission parameters, showed improved robustness to such changes, though it did not match MCTS under the strictest fuel constraint. In contrast, MCTS maintained strong performance in changed-constraint scenarios due to its online re-planning capability, but required over four minutes per test case on average.

These results highlight a fundamental trade-off: learned policies offer rapid inference but may be brittle to unseen constraints, whereas search-based methods provide adaptability at a high computational cost. The inclusion of domain randomization during PPO training partially mitigates this brittleness, suggesting that combining training-time diversity with online adaptability could yield a more balanced solution. Notably, the domain-randomized PPO required almost five times the training steps of the nominal PPO to converge.

A promising direction lies in integrated frameworks such as AlphaZero~\cite{silver2018general} or MuZero~\cite{schrittwieser2020mastering}, which blend neural policy/value predictions with tree search. Such approaches could deliver planners that exploit the speed of learned inference while retaining the flexibility of online search—critical for future ADR missions involving mid-mission refueling, evolving constraints, and uncertain orbital environments.

Future work will focus on:
\begin{itemize}
    \item Investigating hybrid learning–planning approaches for debris rendezvous,
    \item Incorporating uncertainty-aware dynamics to handle orbital perturbations and estimation errors,
    \item Extending to multi-agent collaborative debris removal,
    \item Evaluating real-time performance under strict onboard compute budgets.
\end{itemize}

By bridging learning and planning, we move toward autonomous ADR systems that are both fast and resilient—essential traits for the next generation of space operations.

\section*{APPENDIX}
\addcontentsline{toc}{section}{APPENDIX}  % Optional: adds to TOC

\subsection*{A. Mission Parameters}
\begin{itemize}
    \item \textbf{Initial Orbit and Refueling Orbit:} 700 km almost circular orbit with 96° inclination.
    \item \textbf{Target Debris:} 50 debris objects per episode, sampled randomly between 700--800 km and 94--98°.
    \item \textbf{Refueling Logic:} Allowed after rendezvousing with at least one debris.
    \item \textbf{Maximum \(\Delta v\):} 3 km/s.
    \item \textbf{Mission Duration:} 7 days (per episode).
    \item \textbf{Episode Termination:} Triggers include all debris visited, fuel exhaustion, or mission time-out.
    \item \textbf{Test Case Variation:} Each scenario consists of 100 test cases with 50 randomly initialized debris.
    \item \textbf{J2 Pertubation:} Assuming no effect of pertubations
    \item \textbf{Domain Randomization for Mission parameters for the PPO randomized :} Mission parameters were randomized at the start of each episode:
    \begin{itemize}
        \item $\Delta v_{\max}$ uniformly sampled from [1, 3.5] km/s,
        \item Mission duration uniformly sampled from [1, 7] days.
    \end{itemize}
    \item \textbf{Safety Ellipse:}  
    \begin{itemize}
        \item Semi-major axis: 1 km  
        \item Aspect ratio (radial : along-track): 3 : 1  
        \item Entry condition: relative velocity $v_{\text{rel}} < 0.05$ m/s   
    \end{itemize}
\end{itemize}

\subsection*{B. Reinforcement Learning Hyperparameters}
\begin{itemize}
    \item \textbf{Algorithm:} Proximal Policy Optimization (Masked PPO).
    \item \textbf{Implementation:} Stable-Baselines3-Contrib.
    \item \textbf{Nominal PPO Training Steps:} 1 million.
    \item \textbf{Domain-Randomized PPO Training Steps:} 5.5 million.
    \item \textbf{Learning Rate:} \( 5 \times 10^{-6} \).
    \item \textbf{Batch Size:} 2048.
    \item \textbf{Action Masking:} Enabled (invalid actions assigned \( -\infty \) logits).
    \item \textbf{Policy Network:} MLP with two hidden layers of size 256.
    \item \textbf{Environment Interface:} Gymnasium-compatible.
\end{itemize}

\subsection*{C. Environment and Evaluation Settings}
\begin{itemize}
    \item \textbf{Astrodynamics Libraries:} Poliastro~\cite{poliastro}, Astropy~\cite{astropy}.
    \item \textbf{Computation Platform:} Apple MacBook Pro M1 Max (64 GB RAM), Python 3.11.
\end{itemize}

\subsection*{D. Plain MCTS Configuration}
\begin{itemize}
    \item \textbf{Algorithm:} Plain Monte Carlo Tree Search using UCT.
    \item \textbf{Exploration Constant:} $c_{\text{uct}} = 1.5$.
    \item \textbf{Simulations per Step:} 200.
    \item \textbf{Rollout Policy:} Uniform random over valid actions.
    \item \textbf{Rollout Depth:} 15.
    \item \textbf{Action Masking:} Valid actions filtered using \texttt{valid\_action\_mask()}.
    
\end{itemize}

\bibliographystyle{IEEEtran}
\bibliography{refs}

\end{document}